# Simultaneous Constraint Exploration and Manipulation in Flexible Unknown Environments

Sam Bhattacharyya, and Nabil Simaan*

*Abstract*—This paper presents an online path planning algorithm for safe autonomous manipulation of a flexibly constrained object in an unknown environment. Methods for real time identification and characterization of perceived flexible constraints and global stiffness are presented. Used in tandem, these methods allow a robot to simultaneously explore, characterize, and manipulate an elastic system safely. Navigation without *a-priori* knowledge of the system is achieved using constraint exploration based on local force and position information. The perceived constraint stiffness is considered at multiple poses along an object's (system) trajectory. Using stiffness eigenvector information, global stiffness behavior is characterized and identified using an atlas of simple mechanical constraints, such as hinges and planar constraints. Validation of these algorithms is carried out by simulation and experimentally. The ability to recognize several common simple mechanical constraints (such as a flexible hinge) in real time, and to subsequently identify relevant screw parameters is demonstrated. These results suggest the feasibility of simultaneous global constrain/stiffness exploration and safe manipulation of flexibly constrained objects. We believe that this approach will eventually enable safe cooperative manipulation in applications such as organ retraction and manipulation during surgery.

## I. INTRODUCTION

Commercial surgical assistance systems continue to gain adoption due to their ability to augment surgeons skills (e.g. dexterity and accuracy). Though significant progress has been made by existing commercial systems, they are almost exclusively teleoperated (i.e. they are *passive* manipulators); thus, they ultimately place the entire burden of safeguarding the anatomy on the surgeon. The increased complexity of future and emerging RSAs [1]–[3] calls for the development of *intelligent* robotic slaves that actively gather information about their environment and participate in aiding the surgeon during the completion of certain surgical subtasks. This approach will allow surgeons to focus on crucial aspects of surgical procedures. An example of an application scenario is the manipulation of a large suspended organ by a multitude of robotic arms during retraction. If the organ constraint is known it is possible to safely coordinate the motion of all arms while allowing the surgeon to command movement of one arm. This paper aims to explore the feasibility of simultaneous blind constraint exploration and manipulation in flexible unstructured environments and to propose algorithms that achieve this goal.

This work was supported by NSF Career grant IIS-1063750.

S. Bhattacharyya and N. Simaan are with the Department of Mechanical Engineering, Vanderbilt University, Nashville, TN 37212, USA (e-mails: {sam.bhattacharyya,nabil.simaan}@vanderbilt.edu).
* corresponding author.

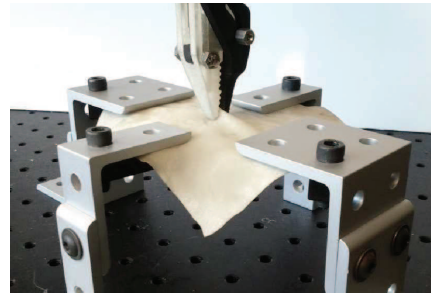

Fig. 1. Membrane constraint, made of a thin sheet of rubber material stretched to moderate tension in a horizontal plane

Some of the first works on path planning in flexible environments were started in the last decade. [4] and [5] present similar methods for modeling deformable objects as a mesh of mass-nodes, inter-connected by spring dampers. In both cases, the path planning dealt primarily with the grasping of flexible objects, in order to maximize the stability of the grasp, while minimizing elastic deformation of the objects being grasped. Using a force based approach and an FEM model, Rodriguez [6] considered the path planning issues of navigating in a completely deformable known environment. By using a Rapidly-Exploring Random Tree algorithm and a collision detection algorithm, they attempt to minimize the elastic energy in all of the possible paths that could be taken to reach the end goal. They used this path planning method to simulate a robot navigating around flexible organs/tissues within the human chest cavity. Patil et al. [7] attempted to automate organ retraction. Like Rodriguez, they assumed a known environment and used FEM to model its elasticity. Their path planning algorithm maximized the exposed tissue area while minimizing maximal tissue stress.

While progress has been made towards modeling of flexible environments and manipulation/grasp in flexible environments, the reliance of FEM methods for surgical applications is a significant hurdle to realistic implementation. Organs are highly inhomogeneous and the act of surgery changes the shape of the environment, thus requiring continuous re-meshing. In this work we take an approach that relies on blind constraint exploration in flexible environments. Our efforts are thus dually focused on the motion planning for safe manipulation in unknown flexible environments, and on exploration of global stiffness properties of an unknown elastic environment.

The contributions of this work lie in its attempts to break new ground in the area of blind constraint exploration and automated manipulation in unstructured flexible environments.

While there have been numerous works on environment exploration (e.g. [8]), stiffness imaging [9], surgical instrument constraint characterization [10], and constraint exploration in rigid environments [11], to the best of our knowledge there are no works dealing with blind constraint characterization and safe manipulation in flexible environments. The paper presents our preliminary efforts to validate the feasibility of simultaneous blind constraint exploration and manipulation in flexible environments. We present methods for exploring and characterizing global stiffness properties of an unknown environment,in real time. Finally, we demonstrate the ability of our algorithm to identify and classify the elastic constraints imposed by several elastic systems.

## II. Nomenclature and Problem Definition

Consider an object suspended in an unknown elastic environment, such as an organ resting within the human body. For a given pose, the *perceived* constraint stiffness matrix is defined as the linear map between an infinitesimal twist $\delta xi$ of the object, and the resulting change in the reaction wrench $\delta \mathbf{w}_e$ from an equilibrium pose.

$$\delta \mathbf{w}_e = \mathbf{K}\delta\xi \tag{1}$$

The perceived constraint stiffness can be modeled as a set of springs acting in parallel on a flexibly suspended object. The local stiffness matrix fully describes the elastic behavior of the object suspension, but only at that pose.

Consider a general 6 DoF gripper which can firmly grasp some portion of the object in a given position/orientation. We will represent the object's pose, $\zeta = [\mathbf{r}^T, \hat{\mathbf{q}}^T]$, using its cartesian position $\mathbf{r}$, and orientation unit quaternion $\hat{\mathbf{q}}$. Likewise, we denote the gripper's velocity in screw coordinates, as a 6-axis twist vector $\xi = [\dot{\mathbf{r}}, \omega]$. While the constrained object might not be perfectly rigid, we assume that the gripper is perfectly rigid and that we have perfect kinematic control over its pose. Finally, we assume that the gripper is equipped with full 6-axis force sensing capability. We will denote the reaction wrench applied on the gripper by the environment by $\mathbf{w_e}$. We also assume that the reaction wrench is entirely due to the elasticity of the environment, and that dynamics does not play a prominent role during quasi-static manipulation.

## III. Motion Planning Algorithm

The task of the online path planner is to manipulate a flexibly suspended object to a target pose while respecting force constraints that prevent damage to the elastic environment. The path planner uses direct sensor data ($\zeta$ and $\mathbf{w_e}$), or localized information that can be deduced from direct sensor data ($\mathbf{K}$ and $E$) to determine the direction of motion at each instant.

### A. Objective Function Definition

The path planner we chose uses the potential field method for its ease of real time computation and flexibility in accepting several optimization constraints. The following mixed reward/penalty objective function is defined:

TABLE I
NOMENCLATURE

| Symbol | Description |
| --- | --- |
| $\xi$ | Instantaneous Manipulator Twist |
| $\mathbf{w}_e$ | Elastic reaction wrench by environment on gripper |
| $\zeta$ | Gripper pose |
| $\mathbf{K}$ | Localized n dimensional stiffness matrix |
| $\tau$ | Spring force in each spring connection to the environment |
| $E$ | Elastic energy of the system, with respect to the initial configuration |
| $J$ | Path Planner Optimization function |
| $\hat{\mathbf{n}}$ | Path Planner Optimal movement direction |
| $\Gamma$ | Diagonal dimensional weighting matrix |
| $^i\Xi_{k-1}$ | The $i^{th}$ stiffness region last identified at stem $k-1$ |
| $^iC_k$ | $i^{th}$ constraint matrix last identified at stem $k$ |

$$J = \sum tasks + \sum constraints \tag{2}$$

where $\sum tasks$ is an attractive potential field that reaches an absolute minimum as the robot satisfies the manipulation tasks and $\sum constraints$ is a penalty (repulsive) potential field that guides the robot away from forbidden manipulation constraints. The online path planner seeks to find a configuration which incrementally minimizes the objective function J.

*1) Task/Reward function:* To achieve manipulation tasks, reward functions $task = f(\zeta, \mathbf{w}_e, \mathbf{K}, E)$ are expressed as a function of the problem variables. For simplicity, we initially chose a basic objective: move the flexibly suspended object to a goal pose $\zeta_g$. This task can be mathematically expressed as the square of the dimensionally weighted distance:

$$task = \zeta_d^T \mathbf{\Gamma}_7 \zeta_d \tag{3}$$

where $\zeta_d$ is the distance from the current to the goal pose, with the first 3 elements as the cartesian distance, and the last 4 elements as the relative orientation $\delta\hat{\mathbf{q}}$ in quaternion space.

$$\zeta_d = [(\mathbf{r}_g - \mathbf{r})^T, \delta\hat{\mathbf{q}}^T]^T \tag{4}$$

As shown in [12], the vector portion of $\delta\hat{\mathbf{q}}$ quantifies orientation error, so we include the 7-dimensional matrix $\mathbf{\Gamma}_7$ to extract and weight the vector portion $\delta\hat{\mathbf{q}}$ by $\alpha$.

$$\delta\hat{\mathbf{q}} = \hat{\mathbf{q}}^{-1} * \hat{\mathbf{q}}_g \tag{5}$$

$$\mathbf{\Gamma}_7 = diag(1,1,1,0,\alpha,\alpha,\alpha) \tag{6}$$

The scalar $\alpha$ is determined by (8) as the ratio of translation to rotation, to produce equivalent amounts of elastic energy. Scalar stiffess $k_x, k_\theta$ are obtained via Frobenius norm of the translational and rotational quadrants of $K$, respectively.

$$E_k = \frac{1}{2}k_x(dx)^2 = \frac{1}{2}*k_\theta*d\theta^2 \tag{7}$$

$$\alpha = \frac{dx}{d\theta} = \sqrt{\frac{k_\theta}{k_x}} \tag{8}$$

*2) Constraint/Cost function:* Like the Task function, the Constraint function requires some manipulation constraint(s) to be defined, and this would normally depend on the specific application. For the purposes of testing our algorithm, we can define two constraints (9) and (10):

$$constraint_1 = \kappa * E \qquad (9)$$

$$constraint_2 = \frac{\rho}{w_{max} - |\mathbf{\Gamma}_6 \mathbf{w}_e|} \qquad (10)$$

where, $\kappa$ is a scaling factor, $W_{max}$ is the maximal allowed magnitude of the force, $\mathbf{\Gamma}_6$ is a 6 dimensional weighting matrix (11), and $\rho$ is a characteristic length, and determines the constraint's radius of influence.

$$\mathbf{\Gamma}_6 = diag([1,1,1,\alpha,\alpha,\alpha]) \qquad (11)$$

The first constraint penalizes for increase in stored elastic energy of the enviroement and the second constraint prevents the manipulator from exceeding a maximal force.

*B. Algorithm*

Since the purpose of the path planner is to find configurations which minimize J, a local optimization routine is sufficient to achieve the given objectives. This yields the core of **Algorithm 1**.

---
**Algorithm 1** Autonomous Manipulation Algorithm
---
**Initialization** at $\zeta_o$
- Evaluate $\mathbf{K} = \frac{\delta \mathbf{w}_e}{\xi dt}$ where dt=0.4 sec
- Set $E = 0$
  **repeat**
    - Measure $\mathbf{w}_e$
    - Direct Kinematics to obtain $\zeta$
    - Numerically evaluate $\nabla J(\zeta, \mathbf{w}_e, \mathbf{K}, E)$
    - Set Manipulator twist $\xi \propto -\nabla J$
    **return** $\xi$
    - Update $\mathbf{K}, E$
  **until** $task(\zeta) < \epsilon$
---

This algorithm is experimentally validated in section V. Despite its apparent simplicity, the algorithm is shown to be effective for using force measurements and direct kinematics to achieve a given manipulation goal.

## IV. METHODS FOR CONSTRAINT IDENTIFICATION

The stiffness matrix $\mathbf{K}$ of a flexibly suspended object is defined as the linear map between an infinitesimal pose perturbation $\delta \xi$, and the resulting reaction wrench $\delta \mathbf{w}_e$ from it's current configuration, Eq. (1). The *local* stiffness at a given pose of the suspended object can be decomposed as a set of springs acting in parallel. This decomposition is not unique [13], [14]. One such decomposition is the eigenscrew decomposition, and it has been shown in ( [15], [16]) to yield the directions of the principal compliant and rigid axes of the elastic constraining environment. These axes can be used to fully describe the form of the mechanical constraint at any configuration of the system. While a local stiffness matrix describes the localized stiffness properties, the information obtained about the stiffness eigenscrews at the current configuration can be used to identify *stiffness regions* that characterize uniformly perceived constraints within a region about a given configuration.

*A. Regional Stiffness*

Consider a given mechanical constraint, such as the gripper in contact with a flexible membrane (Fig. 1). We propose the following:

*Axiom 1:* Stiffness Region

- For a given system in which there is mechanical contact or coupling, which produces a specific mechanical constraint as specified by stiffness eigenscrews, there can be multiple configurations in which this coupling also exists and produces approximately the same basic mechanical constraint
- We define a 'Stiffness region' $\Xi$ as the set of configurations in which a specific mechanical coupling exists, and yields the same basic mechanical constraint

While the gripper of Fig. 1 remains in contact with the flexible membrane, it is subject a 'membrane' constraint. While this constraint is active, the gripper faces impedance in the direction normal to the plane. If the gripper loses contact with the membrane, the coupling will no long exist, and the gripper will not be subject to the same set of mechanical constraints. The stiffness regions of these constraints are the configurations of the gripper for which those constraints are active. As the gripper moves from one stiffness region to another, the former constraints do not 'disappear', but rather, they are just inactive. The gripper will once again be subject to those prior constraints if it moves back to the initial stiffness region.

The stiffness region is accordingly defined as the set of all configurations which share exhibit the same mechanical constraint. The global stiffness of an environment would then be composed not of an infinite number of local stiffnesses, but rather a finite number of identifiable stiffness regions. Mathematically, a stiffness region is expressed as:

$$\Xi \equiv \{\zeta | \zeta \Leftrightarrow C\} \qquad (12)$$

where C is a mechanical constraint, which is itself defined in (13) by a set of screw springs, **e**, as obtained from eigenscrew decomposition of a local stiffness matrix $\mathbf{K}$. By the definition, the stiffness region $\Xi$ is inherently coupled to the defining constraint, i.e. $\Xi \Leftrightarrow C$.

$$C \equiv \{\mathbf{e}_1, \mathbf{e}_2 ... \mathbf{e}_i\} \qquad (13)$$

We hence treat C as a property of a stiffness region, and use it for identifying constraint types and defining stiffness regions.

TABLE II
HEURISTIC CONSTRAINT AXES CLASSIFICATION

|  | $h < \gamma_\theta$ | $\gamma_\theta < h < \gamma_x$ | $h > \gamma_x$ |
|---|---|---|---|
| $\lambda < \gamma_c$ | Free Rotation | Free Screw | Free translation |
| $\gamma_c < \lambda < \gamma_r$ | Torsion Spring | Screw Spring | Linear Spring |
| $\lambda > \gamma_r$ | Rigid, Rotational | Rigid, Screw | Rigid, Translational |

### B. Global Stiffness Exploration

We propose the following constraint exploration algorithm to be executed concurrently with the motion control of the robot. This algorithm continuously discovers and defines new stiffness regions as a robot moves through or explores the workspace. By accumulating a set of known stiffness regions, a rough model of the global environment is developed. This approximate model provides valuable information for use in any future path planning/manipulation within the environment. Furthermore, if a general elastic model already exists for the unknown environment, stiffness regions can be used for stiffness-based localization and registration.

### C. Constraint Identification and Classification

We define perceived mechanical constraints $C$ here as a set of spring-screws, $\{\mathbf{e}_1, \mathbf{e}_2 ... \mathbf{e}_6\}$, obtained from the eigenscrew decomposition of $\mathbf{K}$. To identify if a previous defined constraint $C_i$ is active, using $\mathbf{K}_k$ at a newly explored configuration, we compare each individual eigenscrews which define $C_i$, with each of the eigenscrews derived from $\mathbf{K}_k$.

$$\mathbf{e}_i \sim \mathbf{e}_k \Leftrightarrow \frac{(\mathbf{e}_i - \mathbf{e}_k)(\mathbf{e}_i - \mathbf{e}_k)^T}{\mathbf{e}_k \mathbf{e}_k^T} < \gamma^2 \quad (14)$$

If the normalized difference (14) is below a threshold $\gamma$ (25%), the eigenscrews are concluded to be similar. If every constraint eigenscrew has a match with a locally derived eigenscrew, the local configuration is matched with the constraint. In the event that matches are not found, for a sufficient number of local configurations, the local stiffness is explicitly re-evaluated and a new $\Xi$ and $C$ are defined.

A heuristic classification of constraints can be useful and provide useful information, should a manipulator actually encounter such an elementary constraint. To do so, we classify each of the individual eigenscrews, in a certain constraint, along two sets of categories, based on the pitch $h$ of the eigenscrews, as well as the magnitude $\lambda$, as shown in in table II. $h$ is obtained from the normalized eigenscrews, shown in (15).

$$h = \frac{1}{2}\hat{\mathbf{e}}_i^T \Delta \hat{\mathbf{e}}_i \quad \Delta = \begin{pmatrix} 0 & I_{3x3} \\ I_{3x3} & 0 \end{pmatrix} \quad (15)$$

In the table, the $\gamma$ values represent critical scalar values which are used to identify whether the axis is rotational vs. translational, or compliant vs. rigid. These constants are application specific, as they depend on the size of the workspace, and the overall flexibility of the environment.

A library-based classification can then be used to recognize common constraints. Each pre-defined library constraint would have a set of rules for detecting this constraint from the list of experimentally measured constraints. For example, a flexible hinge constraint would have one compliant torsion axis, and a compliant translational axis perpendicular to it. Algorithm 3, below, details the detection rules used to identify constraints in the following sections.

---

**Algorithm 2** Global Stiffness Exploration

**Initialization at** $\zeta_o$
 - Explore and evaluate the local stiffness $\mathbf{K}_o$
 - Eigenscrew decomposition: $\lambda_o \mathbf{e}_o = \mathbf{K}_o \mathbf{e}_o$
 - Define and classify constraint $^1C \equiv \{\mathbf{e}_1, \mathbf{e}_2 ... \mathbf{e}_i\}$
 - Define new stiffness region $^1\Xi_o = \{\zeta_o\}$, $^1\Xi_o \Leftrightarrow {}^1C$
**while** Exploring **do**
 **At current configuration** $\zeta_k$
 - Eigenscrew Decomposition: $\lambda_k \mathbf{e}_k = \mathbf{K}_k \mathbf{e}_k$
 - Compare eigenscrews to the n known Stiffness regions
 **if** $\{\mathbf{e}_{k,1}, \mathbf{e}_{k,2}...\} \approx {}^iC, i = 1, 2, ...n$ **then**
  - Update $^i\Xi_k = \{^i\Xi_{k-1}, \zeta_k\}$
  - Update $^iC_k = f(^iC_{k-1}, \{\mathbf{e}_{k,1}, \mathbf{e}_{k,2}...\})$
 **else**
  -Define and classify constraint $^kC \equiv \{\mathbf{e}_{k,1}, \mathbf{e}_{k,2}...\}$
  -Define new stiffness region $^{n+1}\Xi_k \equiv \{\zeta_k\}$
 **end if**
 -Move to next configuration $\zeta_{k+1}$
 -Update $\mathbf{K}$
**end while**

---

**Algorithm 3** Heuristic Constraint Identification Algorithm

*Given* $R = {}^r\mathbf{e}_1, {}^r\mathbf{e}_2...$ the set of all rotational axes
and $T = {}^t\mathbf{e}_1, {}^t\mathbf{e}_2...$ the set of all translation axes
**Flexible Hinge Conditions**
${}^r\mathbf{e}_{min} = min(R) \ll avg(R)$ **and** ${}^t\mathbf{e}_{min} = min(T) \ll avg(T)$
${}^r\mathbf{e}_{min} \perp {}^t\mathbf{e}_{min}$
Resulting lever arm $r_l$:  $r_{min} < r_l < r_{max}$
**Linear Spring Constraint Conditions**
${}^t\mathbf{e}_{max} = max(T) \gg avg(T)$
$|{}^t\mathbf{e}_1| \approx |{}^t\mathbf{e}_2| \ll |{}^t\mathbf{e}_{max}|$
**Membrane Constraint Conditions**
${}^t\mathbf{e}_{min} = min(T) \ll avg(T)$
$|{}^t\mathbf{e}_1| \approx |{}^t\mathbf{e}_2| \gg |{}^t\mathbf{e}_{min}|$

---

## V. EXPERIMENTAL SETUP

The blind path planning algorithm was first tested on a planar test case, whereby a movable rigid triangle was attached by 3 springs in parallel to a fixed base platform, as shown in shown in Fig. 2. The goal of the experiment was to validate the ability of the motion planning algorithm to move the triangle from it's initial equilibrium pose a goal pose while enforcing a pre-defined set of constraints on the maximal elastic force on each constraining spring. The experimental setup used a Puma560 robot controlled using Matlab xPC Target real time operating system with an ATI Gamma six axis force sensor and a custom made gripper. The robot was controlled with a 1KHz control loop and the force sensing data was filtered using a 200 point moving average filter.

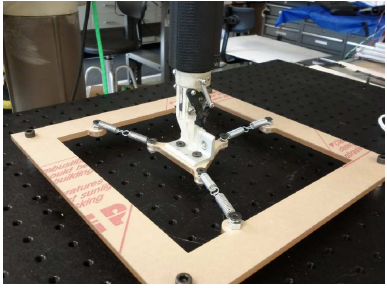

Fig. 2. Experimental setup, involving a Puma560 robot with a force sensor and gripper attached. A flexibly suspended triangular platform is safely manipulated to a target configuration

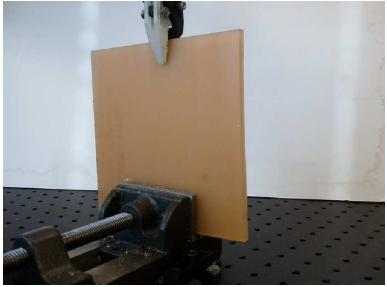

Fig. 3. A real flexible hinge constraint, using a 1/4" thick Polyeurethane sheet attached at it's base to a clamp

The constraint identification (exploration) algorithm was evaluated on three other experimental setups shown in Fig. 4, Fig. 3 and Fig. 1. The setup in Fig. 4 represents an object constrained by a line spring. The setup in Fig. 3 uses a flexible polyeurethane plate that is gripped between the jaws of a vice; thus, trying to mimic a flexible hinge constraint.

## VI. RESULTS

### A. Path planning

Given a target pose, the autonomous manipulation algorithm was able to move the rigid triangle to the goal pose, or get as close as possible without violating the constraints. While algorithm 1 is elementary, it is important to emphasize it's effectiveness and repeatability. The results from two such runs are shown in Fig. 5 and Fig. 6. In one of the runs, the goal pose is $[.025, .02, 0]^T$, while the other has a goal pose of $[.035, .04, .0078]^T$. These poses are listed in planar coordinates $([x, y, \theta]^T)$ in units of [m, m, rad], respectively. In the latter run, the robot is prevented from moving to the goal pose to avoid violating a maximal spring force constraint of 25 N. In this way, the constraint function acts like a basic virtual fixture, however it operates in the unknown force domain, as opposed to the spatial domain.

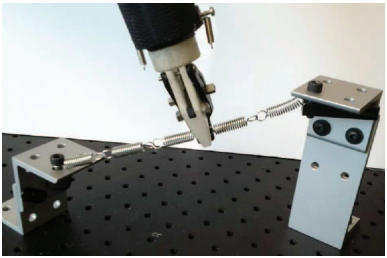

Fig. 4. Linear Spring constraint, created by a series of linear springs, with two fixed ends in space

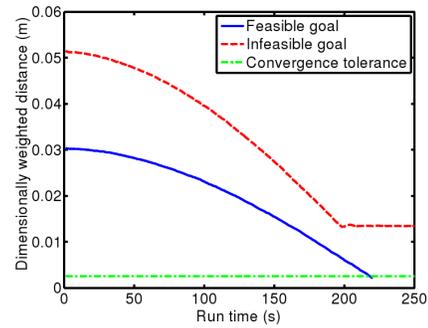

Fig. 5. Position error during autonomous navigation experiment

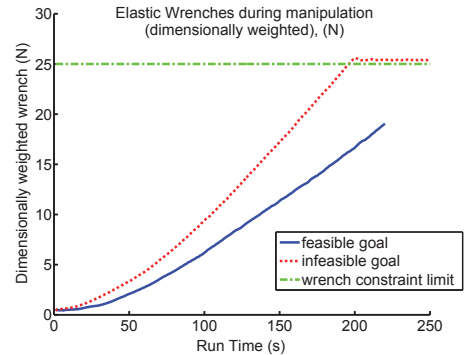

Fig. 6. Normalized elastic wrench during autonomous navigation experiments

### B. Exploration and constraint identification

In all three constraint identification experiments, Algorithm 3 correctly identified the intended constraint. For the line spring setup in Fig. 4, the eigenvectors, eigenvalues and screw-pitches are given in(16,17, 18). The bold eigenvector/eigenvalue corresponds to the stiffest axis, which is translational ($h \approx 0$), and roughly parallel to the direction of the spring (19).

$$\begin{pmatrix} \mathbf{-0.648} & -0.516 & -0.559 & -0.002 & 0.002 & -0.000 \\ \mathbf{-0.734} & 0.230 & 0.637 & -0.011 & -0.027 & 0.009 \\ \mathbf{0.200} & -0.824 & 0.528 & -0.002 & -0.009 & 0.019 \\ \mathbf{0.018} & -0.000 & -0.024 & -0.197 & -0.977 & -0.066 \\ \mathbf{-0.004} & -0.013 & 0.018 & 0.140 & 0.038 & -0.989 \\ \mathbf{-0.005} & 0.000 & 0.000 & 0.970 & -0.204 & 0.129 \end{pmatrix} \quad (16)$$

$$(\mathbf{1617.0},\ 789.1,\ 650.4,\ 1.315,\ 0.611,\ 1.017) \quad (17)$$

$$(\mathbf{0.0094}\ \ 0.0036\ \ 0.0253\ \ 29.1843\ \ 0.8828\ \ 14.9692) \quad (18)$$

$$\begin{pmatrix} 0.1500 & 0.1000 & 0.0710 \end{pmatrix} \qquad (19)$$

The second set of experiments carried out using the setup in Fig. 3 produced the results shown in Eqs. (20,21,22). Equation (20) presents the wrenches associated with the eigenscrews. The two bold screws show the most compliant translational and rotational axes. The intended hinge in Fig. 3 was aligned with the x axis of the robot. The results show that the most compliant axis is a vector in the xz diagonal. This possibly incorrect result could be due to a misalignment of the gripper, bending the edge of the relatively stiff 6.5 mm Polyurethane sheet.

$$\begin{pmatrix} 0.126 & 0.950 & -\mathbf{0.283} & 0.018 & -0.013 & \mathbf{0.011} \\ -0.013 & 0.288 & \mathbf{0.955} & -0.005 & 0.045 & -\mathbf{0.031} \\ 0.991 & -0.117 & \mathbf{0.048} & 0.012 & 0.026 & \mathbf{0.008} \\ -0.016 & -0.011 & \mathbf{0.031} & 0.752 & -0.051 & \mathbf{0.655} \\ 0.024 & -0.001 & \mathbf{0.048} & -0.076 & -0.995 & \mathbf{0.007} \\ 0.000 & 0.008 & \mathbf{0.015} & -0.654 & 0.056 & \mathbf{0.753} \end{pmatrix} \qquad (20)$$

$$\begin{pmatrix} 1671.0, & 1142.0, & \mathbf{200.3}, & 1.211, & -1.779, & \mathbf{0.344} \end{pmatrix} \qquad (21)$$

$$\begin{pmatrix} 0.0019 & 0.0130 & \mathbf{0.0383} & 23.7914 & 14.6042 & 7.4824 \end{pmatrix} \qquad (22)$$

The third set of experiments was carried out on the membrane constraint setup shown in Fig. 1, resulting in the eigenscrews, eigenvalues and pitches in (23, 24, 25). The membrane was 0.5 mm thick made of Latex rubber laid out in the horizontal x-y plane. The eigenvalues clearly show that the least stiff direction is in the z axis (indicated by a boldface column) and that the rotational stiffness of the membrane is negligible.

$$\begin{pmatrix} 0.275 & 0.922 & -\mathbf{0.269} & 0.000 & 0.022 & -0.011 \\ 0.958 & -0.285 & \mathbf{0.003} & -0.006 & -0.003 & -0.004 \\ 0.073 & 0.260 & \mathbf{0.961} & -0.005 & -0.038 & 0.013 \\ -0.008 & 0.001 & -\mathbf{0.010} & -0.837 & -0.276 & -0.444 \\ -0.003 & -0.012 & \mathbf{0.043} & -0.037 & 0.888 & -0.533 \\ -0.000 & 0.000 & -\mathbf{0.004} & 0.545 & -0.362 & -0.718 \end{pmatrix} \qquad (23)$$

$$\begin{pmatrix} 1188.0, & 996.9, & \mathbf{365.2}, & 0.660, & -3.250, & 4.039 \end{pmatrix} \qquad (24)$$

$$\begin{pmatrix} 0.0055 & 0.0050 & \mathbf{0.0575} & 38.8509 & 2.5722 & 34.9887 \end{pmatrix} \qquad (25)$$

## VII. CONCLUSION

This paper proposed algorithms for blind exploration and safe manipulation and constraint identification in flexible unstructured environments. Using non-spatial cost functions for path planning, a computationally efficient motion planner was used to safely manipulate an elastically constrained objects. Using local stiffness and eigenscrew analysis to characterize local global elastic behavior, it was shown that environmental elastic properties could be identified and characterized in real time. The validation methods used in this paper are fundamental, but elementary. Future work in this direction would include developing more comprehensive, intelligent and robust constraint identification methods, using local stiffness and eigenscrews. Furthermore, it would include seamlessly integrating and implementing the path planning, global stiffness and constraint identification algorithms, for execution of real surgical tasks in an unknown environment. These developments are crucial, as they represent the early steps on the road to automation, not only for RSAs, but for any robot which operates in unknown elastic environments.